\documentclass[namedreferences,hyperref,optionalrh]{spr-sola}
\usepackage{graphicx}        
\usepackage{amssymb}        
\usepackage{color}           
\usepackage{breakurl}                         

\newcommand{\R}{\mathbb{R}}
\newcommand{\E}{\mathcal{E}}
\newcommand{\C}{\mathcal{C}}
\newcommand{\F}{\mathcal{F}}
\renewcommand{\S}{\mathcal{S}}
\renewcommand{\L}{\mathcal{L}}

\chardef\us=`\_

\begin{document}

\begin{frontmatter}

\title{Machine learning for reconstruction of polarity inversion lines from solar filaments}

\author[addressref={aff1},email={waclove@yandex.ru}]{\inits{V.}\fnm{Vaclovas}~\lnm{Kisielius}\orcid{0000-0002-3204-9031}}

\author[addressref={aff1,aff2},email={egor.illarionov@math.msu.ru}]{\inits{E.A.}\fnm{Egor}~\lnm{Illarionov}\orcid{0000-0002-2858-9625}}

\address[id=aff1]{Moscow State University, Moscow, Russia}
\address[id=aff2]{Institute of Continuous Media Mechanics, Perm, Russia}

\runningauthor{Kisielius \& Illarionov}
\runningtitle{\textit{Solar Physics} Reconstruction of polarity inversion lines}

\begin{abstract}
    Solar filaments are well-known tracers of polarity inversion lines 
    that separate two opposite magnetic polarities on the solar photosphere.
    Because observations of filaments began long before the systematic observations of solar magnetic fields, historical filament catalogs can facilitate the reconstruction of magnetic polarity maps at times when direct magnetic observations were not yet available. In practice, this reconstruction is often ambiguous and typically performed manually. We propose an automatic approach based on a machine-learning model that generates a variety of magnetic polarity maps consistent with filament observations. To evaluate the model and discuss the results we use the catalog of solar filaments and polarity maps compiled by McIntosh. We realize that the process of manual compilation of polarity maps includes not only information on filaments, but also a large amount of prior information, which is difficult to formalize. In order to compensate for the lack of prior knowledge for the machine-learning model, we provide it with polarity information at several reference points. We demonstrate that this process, which can be considered as the user-guided reconstruction or super-resolution, leads to polarity maps that are reasonably close to hand-drawn ones, and additionally allows for uncertainty estimation.
\end{abstract}
\end{frontmatter}

\section{Introduction}
    About 50 years ago, deriving magnetic-field information from chromospheric observations was the main way to compensate for the lack of magnetic-field maps required for practical purposes and research. Patrick McIntosh, one of the founders of the space weather field, mentioned in \cite{McIntosh1972} up to ten advantages of this approach that were actual for those days. Most of these advantages are less relevant today due to the availability of full-disk solar magnetographs with high spatial and temporal resolution. However, at least one of these advantages remains actual -- the existence of chromospheric observations for more than 120 years 
    \citep[at the time, McIntosh said 70 years; for an extended time period, see, e.g.,][]{Tlatova, Mazumder_2021, Chatzistergos}.
    This motivates continuing attempts to use these data to reconstruct magnetic-field maps.

    The approach described in \citet{McIntosh1976} focuses on the reconstruction of magnetic polarity only. He considers the five basic chromospheric structures (filaments, filament channels, fibrils, arch-filament systems, and plage corridors) derived from H$\alpha$ observations and uses their positions to infer the polarity inversion lines, which in turn define the magnetic polarity map. It should be realized that the approach is not an exact algorithm, but an analysis, combined with the author's experience and understanding of what can and cannot be. For this reason, the dataset created by \cite{mcintosh_archive}, which spans about 45 years, remains a unique resource in the field of solar physics.
    
    The downside of this unique approach is that it can unlikely be consistently extrapolated forward or backward in time. Such attempts were made in \cite{Makarov_partI} and \cite{Makarov_partII} and cover the time period from 1904 to 1981 based on Kodaikanal, Meudon, and later H$\alpha$ observations. However, for many reasons, this Archive should also be considered as a unique and non-reproducible approach.

    This motivates us to investigate an algorithmic way for reconstruction of magnetic polarity. Keeping in mind the idea of inferring magnetic polarity in historical period, we restrict the problem to the most prominent tracer of polarity inversion lines -- solar filaments. These solar features tend to lie on the polarity inversion line 
    (\citealp[see, e.g.,][]{McIntosh1972, Makarov_partI}; \citealp[and more recent][]{Durrant2002,Ipson2005,Mazumder_2018}), and the problem of magnetic polarity reconstruction can be formulated as the reconstruction of polarity inversion lines based on fragments of these lines. We realize that this formulation is not strictly equivalent to that considered, for example, by McIntosh or Makarov, but it is probably the most straightforward for an algorithmic approach. The detailed formalization will be given below.
    
    Although we are only interested in the reconstruction of the magnetic polarity, there were attempts to derive the intensity of the magnetic field from chromospheric H$\alpha$ observations. The first systematic approach, apparently, was proposed in \cite{Veeder}, and it was recently revisited using a deep learning model in \cite{Gao2023}. Note, however, that the latter approaches use full-disk spectrograms, whereas we will proceed from much more scarce information on the shape and location of the filaments.
    
    As we discuss later, the main problem for the algorithmic approach is the proper mathematical formalization of the reconstruction process. Once this is done, the solution to the mathematical problem can be found using various methods. In our research we will use the machine-learning approach based on a neural network model. One more point is that any reconstruction should be considered together with the corresponding uncertainty estimates. We will demonstrate that the proposed approach naturally provides such estimates by generating a set of possible reconstructions.

    The rest of the paper is organized as follows. Section 2 presents the synoptic maps derived from the Patrick McIntosh Archive and discusses the preprocessing steps taken to prepare the data for analysis. Section 3 describes the architecture of the neural network model and the loss function used during training. In Section 4, we present the results of polarity reconstruction and discuss them in Section 5.

\section{Data}
\subsection{McIntosh Archive of Synoptic Maps}
    We use a digitized version of the McIntosh Archive of Synoptic Maps, as referenced in \cite{mcintosh_archive}. This dataset contains a continuous series of synoptic maps for Carrington rotation (CR) numbers from 1355 to 2086 that cover the time period from 1954 to 2009. For our test study we will use only the first 36 maps (i.e., CRs from 1355 to 1380). The digitized synoptic maps have a resolution of $2009 \times 4018$ pixels, and each pixel encodes information about the polarity of the magnetic field, the polarity of coronal holes, the presence of filaments, sunspots, plages, and polarity inversion lines in that pixel.
    
\begin{figure}
        \centering
        \includegraphics[width=0.6\textwidth]{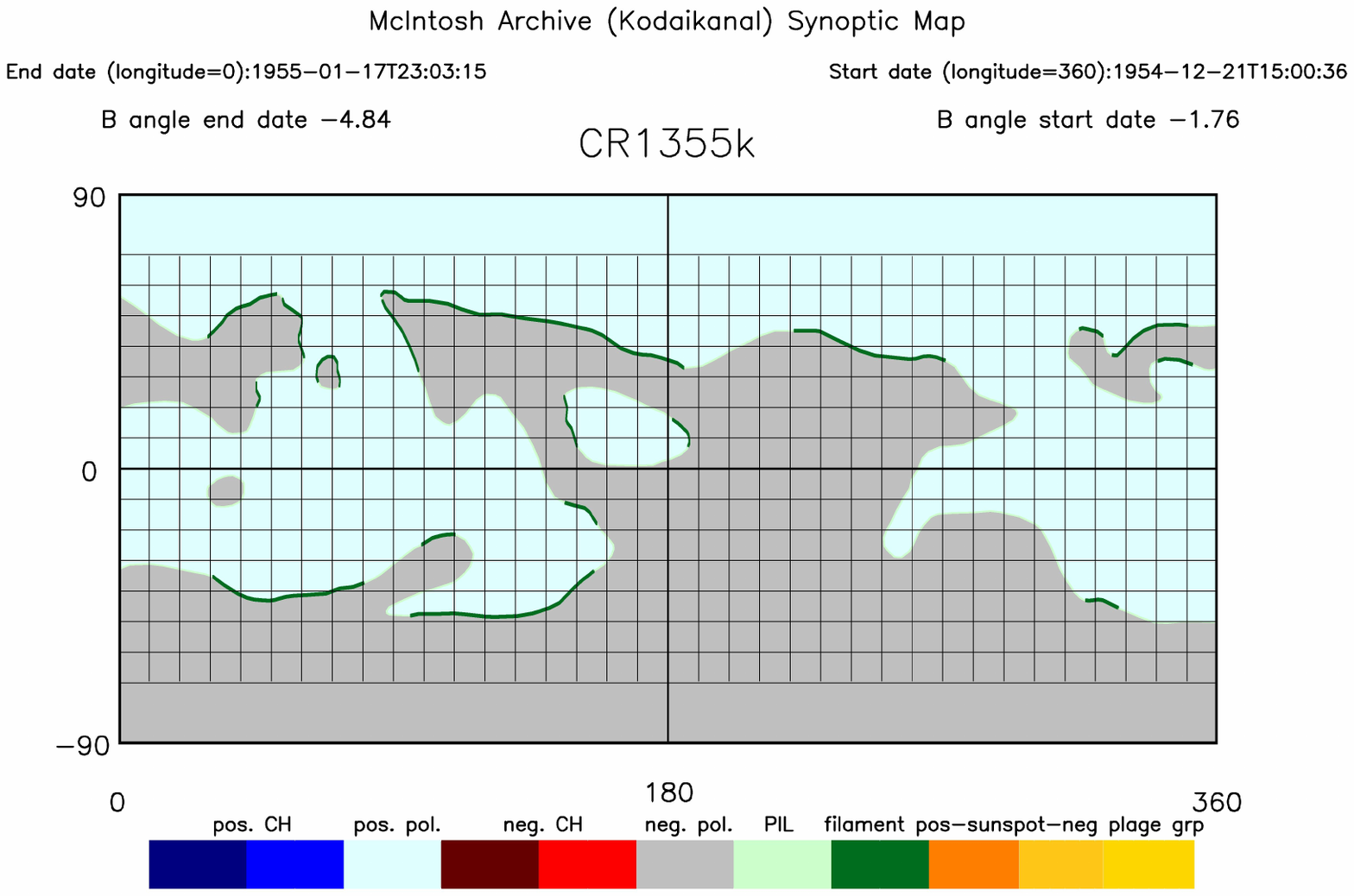}
        \caption{Example of a synoptic map from McIntosh Archive.}
        \label{fig:mcintosh-map}
    \end{figure}

    Figure~\ref{fig:mcintosh-map} shows a map from this dataset. In our research, we will use information only about filaments and polarity of the magnetic field.
    
\subsection{Preprocessing}
    The original 4K resolution images from the McIntosh Archive were reduced  to $256\times512$ pixels to reduce computational costs.
    
\begin{figure}
        \centering
        \includegraphics[width=0.75\textwidth]{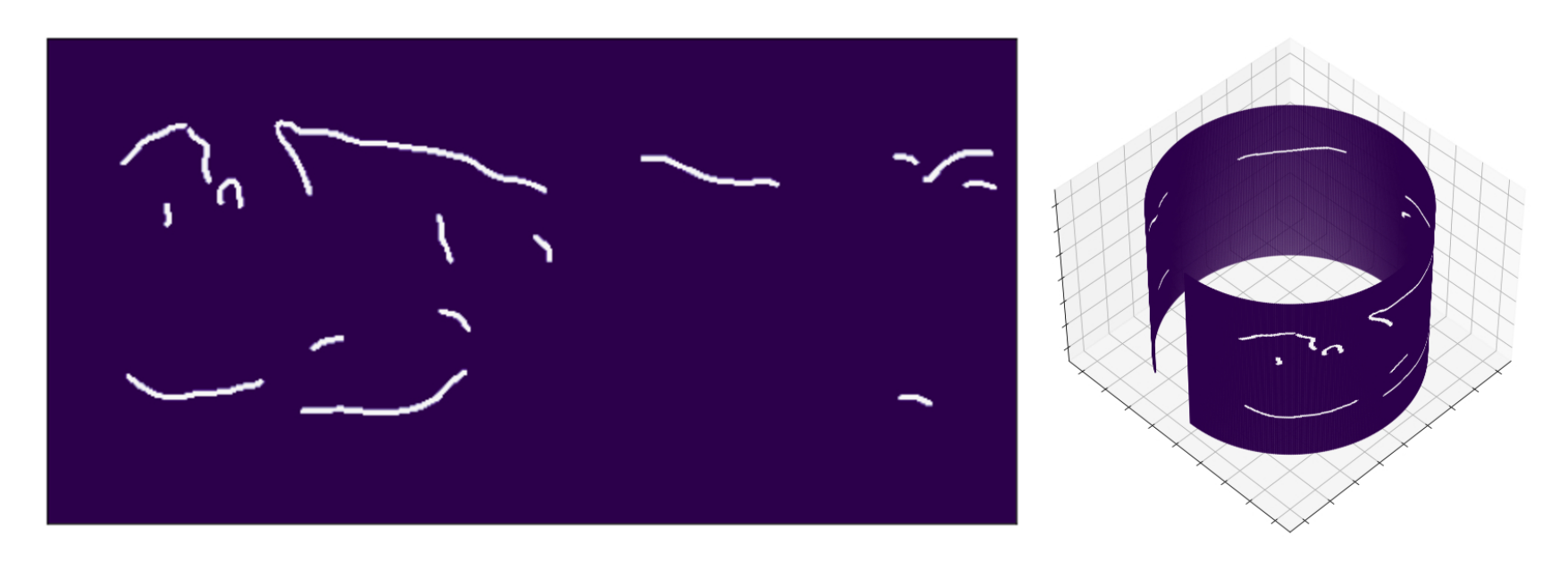}
        \caption{Projection of a synoptic map of solar filaments from 2D geometry (the left panel) onto cylinder $\C \subset \R^3$ (the right panel). The cylindrical projection contains a small gap between the edges of the image. This gap was introduced to account for the fact that there is no exact match between the left and right edges of the image, but only a similarity.}
        \label{fig:maptocyl}
    \end{figure}

    Although the images (synoptic maps) are given in plane geometry,
    they are actually a fold out of the solar surface. The left and right edges of these images are not identical due to the magnetic field changing over time, but they are quite similar because the magnetic field change is relatively slow compared to the Sun's rotation period. We take this feature into account by projecting these images onto the surface of a cylinder $\C \subset \R^3$. This means that each pixel now has three coordinates $x$, $y$ and $z$ that encode its position in cylindrical projection in 3D space. Figure~\ref{fig:maptocyl} shows an example of a synoptic map encoding position of filaments in original 2D geometry and the corresponding mapping to a cylinder. Looking ahead, we note that this cylindrical mapping has proven to be more efficient than the original 2D geometry and also more efficient than the mapping on a sphere.
    
    As a result of this step, we obtain three arrays for each synoptic map. The first array contains 3D coordinates of pixels of the synoptic map. The second array contains for each pixel a value of 0 or 1, where 1 means that this pixel is part of the filament and 0 means that it is not. The third array contains for each pixel a value of $-1$ or 1, where $-1$ means negative polarity and 1 means positive polarity for that pixel.

\section{Model}
\subsection{Mathematical formalization}
    We proceed from the simplification that filaments are features on the solar photosphere, which is considered as a unit sphere $\S$ in three-dimensional space. Each point on the sphere can be defined by a radius vector $\vec{r}$. Then we consider the polarity map as a field defined by some function $f(\vec{r}): \S \to [-1, 1]$, and the polarity inversion line (PIL) as the zero-level of that function, i.e. PIL is a set $\{\vec{r}: f(\vec{r})=0 \}$. Correspondingly, any negative values of the function are interpreted as the negative polarity, positive values of the function -- as the positive polarity.

\begin{figure}
        \centering
        \includegraphics[width=0.75\textwidth]{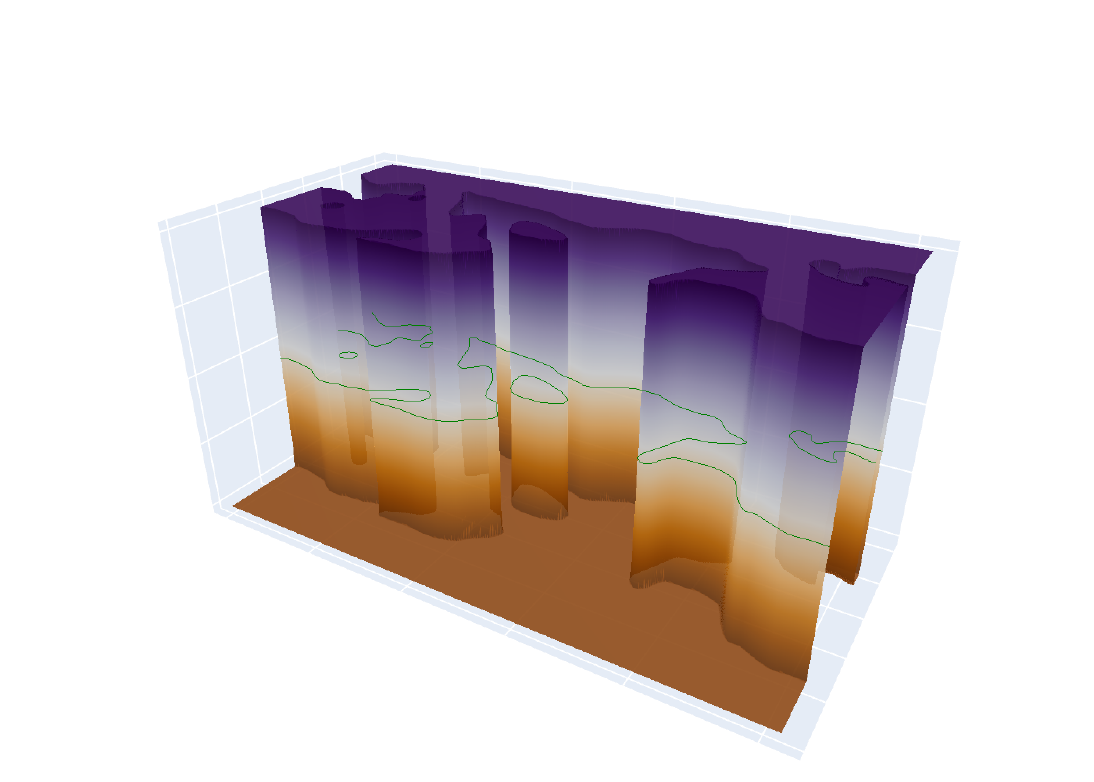}
        \caption{Illustration of the function $f(\vec{r})$ defined on the plane of the synoptic map and approximating the polarity map. The orange color is for negative values of the function, the purple color is for positive values of the function. The green line is the zero-level
        of the function and is interpreted as the polarity inversion line.}
        \label{fig:target3d}
    \end{figure}

    Then we consider two projections of the sphere -- on a plane, which yields the synoptic map, and on a cylinder, as it was shown in ~\ref{fig:maptocyl}. Again, for simplicity, we neglect projection issues near the poles since these regions are essentially unipolar except for the reconnection periods. Figure~\ref{fig:target3d} shows an example of the function $f(\vec{r})$ considered for the plane projection.

    The observational data (filaments) provide information about some fragments of the PIL, i.e., a subset of points $\vec{r}$ at which $f(\vec{r})=0$ is known. The idea of the reconstruction is to find a function $f$ consistent with these observational constraints. We parameterize the function $f$ with a neural network model and use the condition $f(\vec{r})=0$ to optimize the model. Additional constraints on $f$ that prevent it from being trivial (i.e., a zero function everywhere) will be discussed below when we construct the loss function for the neural network model.

    In the following, we will use the cylindrical projection to solve the reconstruction problem and show the results in projection onto the plane.
    
\subsection{Architecture}
\begin{figure}
        \centering
        \includegraphics[width=\textwidth]{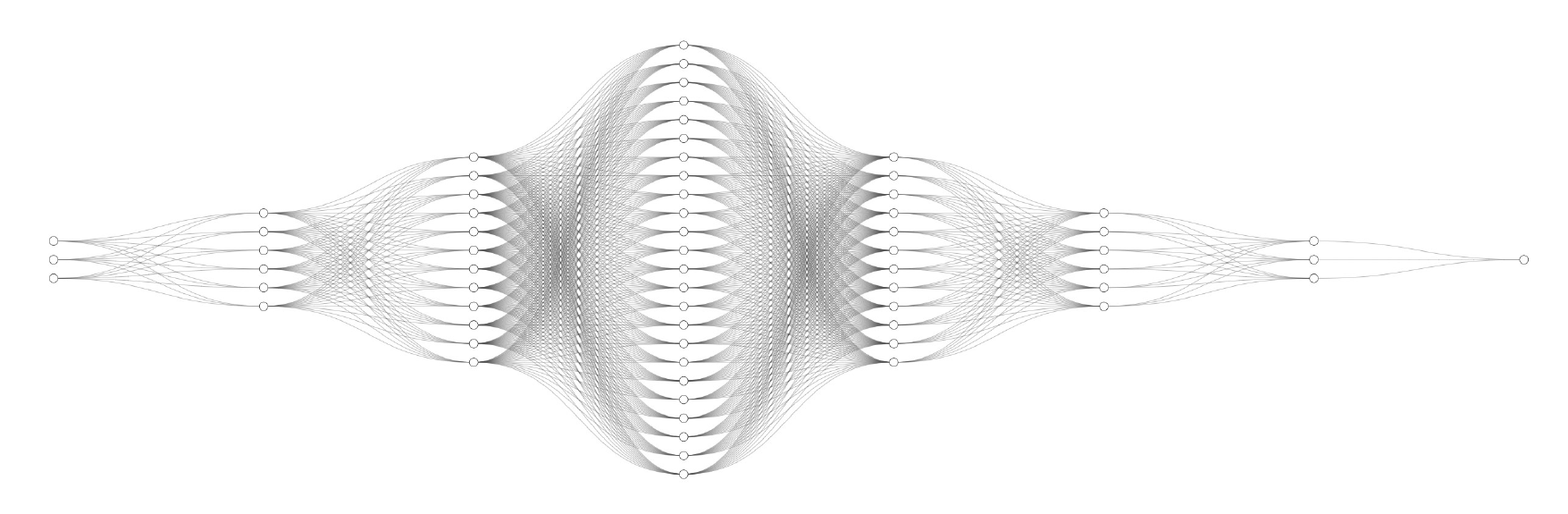}
        \caption{The architecture of the fully connected neural network applied for 
        reconstruction of magnetic polarity maps. Circles represent neurons, lines show
        connections between neurons.}
        \label{fig:nn-architecture}
    \end{figure}

    We approximate the function $f(\vec{r})=0$ with a fully connected neural network model that has three inputs ($x$, $y$, $z$) and one output, followed by a tangent activation function that ensures that the output is in the range $[-1, 1]$. 
    Given a sufficient number of intermediate neurons, such a model can approximate any continuous function $f$  \citep[see][]{Cybenko}.
    After a number of experiments, we arrived at a fully connected architecture consisting of 8 layers with the number of neurons in the layers (3, 6, 12, 24, 12, 6, 3, 1).
    Figure~\ref{fig:nn-architecture} shows a schematic representation of the neural network architecture.
    In total, this model has only 823 trainable parameters, which is very small compared to many other neural network architectures. We will discuss the motivations for using fully connected models and their computational performance in the last section.
    
\subsection{Loss function}
    To train the neural network model we construct a specific loss function $\L(f)$ that accounts for observational constraints (positions of filaments) and a number of
    additional physics-based heuristics. Let's discuss the components of the loss function:
    \begin{itemize}
        \item solar filaments typically align along polarity inversion lines, so the function $f$, defined by the neural network, should have a value of zero in regions corresponding to filaments, denoted as $\F$. We can achieve this by minimizing the integral
        $$
        \int_{\F} f^2 \,dV
        $$
        \item outside the polarity inversion lines, the magnetic field polarity is either positive or negative, implying that $f$ should take values close to $+1$ or $-1$ in the area $\C \setminus \F$. This can be achieved by maximizing the integral
        $$
        \int_{\C \setminus \F} |f| \, dV
        $$
        \item the total magnetic field of the Sun is supposed to be neutral, which means that we need to minimize the integral
        $$
        \left|\int_{\C} f \,dV \right|
        $$
        \item it can be noted that if the function $f$ minimizes the above integrals, then the function $-f$ does as well. To resolve this ambiguity, the polarity of at least one point should be known. It is reasonable to assume that the polarity at solar poles is known, given that magnetic field reversal occurs infrequently and can be estimated. Practically, we prescribe the dominant polarity ($p_N$ for the North pole and $p_S$ for the South pole) to the
        horizontal bands of 10 degrees from the poles and minimize the term 
        $$
        P=\int_{\theta > 80^{\circ}} (f-p_N)^2 \, dV + \int_{\theta < -80^{\circ}} (f-p_S)^2 \, dV \,.
        $$
        
    \end{itemize}
    At this moment, the loss function is of the form:
    \[
        \L(f) =  w_1\left|\int_{\C} f \,dV \right| + w_2 \int_{\F} f^2 \,dV  
        -w_3\int_{\C \setminus \F} |f| \, dV
        + w_4 P \, ,
    \]
    where $w_i$ are normalization weights, inversely proportional to the maximum possible value of the corresponding integrals. 
    
    The above loss function formally requires only that $f$ be close to zero in the region corresponding to the filament, but does not require that the function have opposite signs on opposite sides of the filament.
    To improve this, one could propose an additional term that maximizes the norm of the gradient of the function at the points of filaments, i.e. maximizing
    $$
    \int_{\F} \|{\nabla f}\|^2 \,dV \,.
    $$
    Although this term can indeed be implemented and used to optimize neural networks, our experiments did not show that it helps to obtain more consistent polarity maps.

    Instead, we assume that the polarity is known at several points on the synoptic map. This prior knowledge can be obtained, say, from manual expert annotation, or from low-resolution magnetograms, or from local measurements of magnetic fields. Such semi-supervised (or guided) reconstruction seems more feasible and, to some extent, more natural. Below we consider how the number of reference points (points with known polarity) affects the reconstruction results.

\begin{figure}[h]
        \centering
        \includegraphics[width=\textwidth]{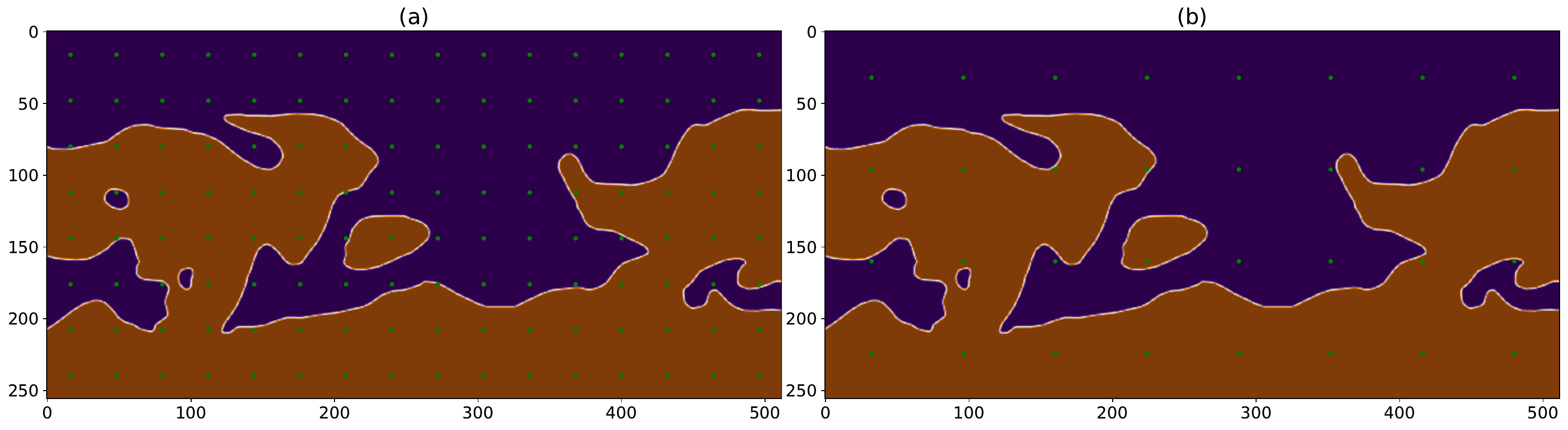}
        \caption{Synoptic maps with auxiliary grids of reference points (green dots). The grid step, indicating the distance between adjacent green dots, is 32 pixels in image (a) and 64 pixels in image (b).}
        \label{fig:ref_points}
    \end{figure}

    We assume that the reference points are defined as a uniform grid on the synoptic map. In our experiments we will vary the step of this grid. Figure~\ref{fig:ref_points} illustrates examples of maps with different steps of the auxiliary grid of reference points. 
    
    To include information from the reference points, we add the MSE term evaluated at the reference points to the loss function. The final loss is
    \[
        \L^\dag(f) = \L(f) + \frac{1}{n}\sum\limits_{i=1}^{n}(f(\vec{r}_i) - p_i)^2 \, ,
    \]   
    where $n$ is the number of reference points, $\vec{r}_i$ and $p_i$ are the coordinates and the polarity ($+1$ or $-1$) of the $i$-th reference point correspondingly.
    
    The model training procedure is to minimize the loss function $\L^\dag(f)$. We use the Adam optimizer \citep{KingBa15} with a learning rate of \(5 \times 10^{-3}\) and a weight decay of \(1 \times 10^{-4}\). The number of iterations is set to $3\times 10^4$, since after that the loss function stops decreasing significantly. Note that for each synoptic map, we use a separate instance of the neural network model initialized with random weights.

\subsection{Confidence estimation}
    By construction, for each point with coordinates ($x$, $y$, $z$) on the synoptic map, the model predicts a value in the range $[-1, 1]$. We interpret the sign as the polarity sign, and the absolute value as the confidence in this sign. However, the results for a single trained model may depend on the initial values of the model parameters and the realization of the stochastic optimization process. To reduce this ambiguity, we train 100 independent models for each synoptic map. This can be interpreted as 100 independent expert opinions for an ill-posed problem.

    To aggregate the multiple predictions into a single map containing only two values of $+1$ or $-1$, we consider two strategies:
    \begin{enumerate}
        \item average the results for each pixel, and then use the zero threshold to assign $+1$ or $-1$;
        \item use a zero threshold to assign $+1$ or $-1$ to each pixel in each of the predicted synoptic maps, and then use the simple majority rule for aggregation.
    \end{enumerate}
    For the first approach, the mean value can be interpreted as confidence in the polarity sign after aggregation.

    In the next section, we demonstrate that the first strategy (using simple averaging) gives the best accuracy. Until then, all plots and numerical results are given for the first aggregation strategy. For simplicity, the use of a zero threshold to assign $+1$ or $-1$ to each pixel will be referred to as binarization.

\section{Results}
    In this section we compare the polarity maps produced by the trained neural network models with polarity maps in the McIntosh Archive. Although we consider polarity maps in the McIntosh Archive as target ones, it should be noted that in the framework of our model any polarity map consistent with the positions of filaments should be considered as a proper one. Therefore, any deviations of the model predictions from the polarity maps in the McIntosh Archive should not be immediately considered as an error. These deviations may indicate variability or uncertainty in the reconstruction results.

    To quantify the difference between the reconstructed polarity map and the corresponding map from the McIntosh Archive, we consider two metrics. The first one, denoted as $\E$, is the fraction of pixels with incorrectly predicted polarity. But since the polar regions are mostly unipolar, we also measure this fraction within the region corresponding to $\pm$40 degrees from the equator \citep[note that the same latitudinal belt was considered in][]{McIntosh1972}. The second metric is denoted as $\E^*$. 

    To demonstrate the results, we will use one synoptic map from the beginning (CR 1355), one from the middle (CR 1373), and one from the end (CR 1387) of the set of synoptic maps. For each of the three selected synoptic maps, we show the input map with filament locations, the target polarity map, the average prediction over 100 realizations of the neural network model (for confidence estimation), and the binarized average prediction (for direct comparison with the target map). For the full set of synoptic maps, we provide a plot with $\E$ and $\E^*$ errors.

\subsection{Grid-Based Reference Points With Step 1}
    We start with a trivial experiment in which we train neural network models on a grid of reference points with step size equal to one, i.e., we give the model full access to the target polarity map during the training phase.
    The goal of this experiment is to demonstrate that the proposed architecture of the
    neural network model is capable to approximate the target polarity map.
    Due to the simplicity of the problem, we train only four independent models for each synoptic map instead of 100.
    
\begin{figure}[h]
        \centering
        \includegraphics[width=\textwidth]{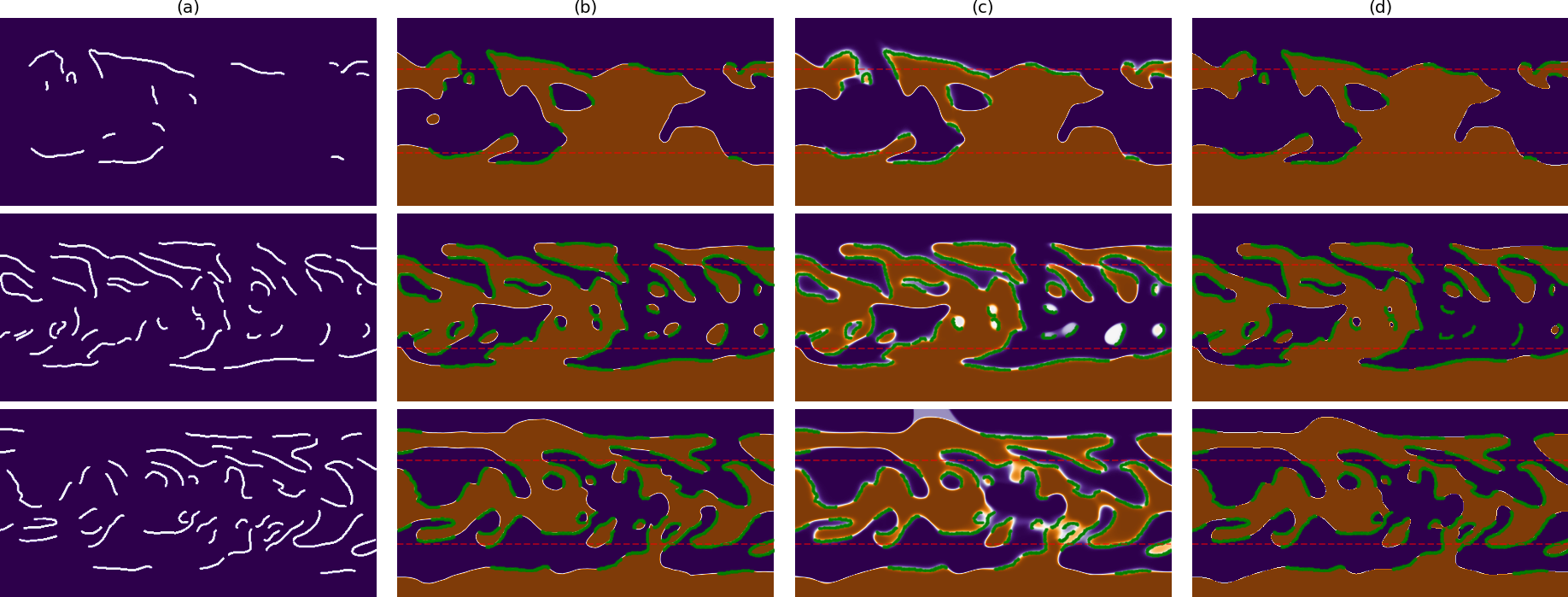}
        \caption{Reconstruction of polarity maps for Carrington rotations 1355 (top row), 1373 (middle row), and 1387 (bottom row). Each row includes four columns: (a) input data with filament structures; (b) target polarity map from the McIntosh Archive; (c) averaged prediction of the models; and (d) binarized averaged prediction. The white lines in column (a) as well as green lines in columns (b)--(d) correspond to filaments. The region within $\pm40$ degrees of latitude around the solar equator where we calculate the $\E^*$ error is marked with red dashed horizontal lines. The grid of reference points has a step size of 1 pixel.}
        \label{fig:example_1}
    \end{figure}

    Figure~\ref{fig:example_1} shows the synoptic maps of filaments, 
    target polarity maps, averaged predictions, and binarized averaged predictions.
    We observe that the model is able to approximate the polarity maps, although not perfectly, and may miss a number of small-scale details. At the same time, we observe that in many cases, the missed regions in the binarized maps reveal as white regions in the maps of averaged predictions. These white regions correspond to averaged predictions close to zero, and we interpret them as regions where confidence in the polarity sign is low. Thus, we see that at least some of the errors are associated with low model confidence in the pixel sign.
    
\begin{figure}[h]
        \centering
        \includegraphics[width=\textwidth]{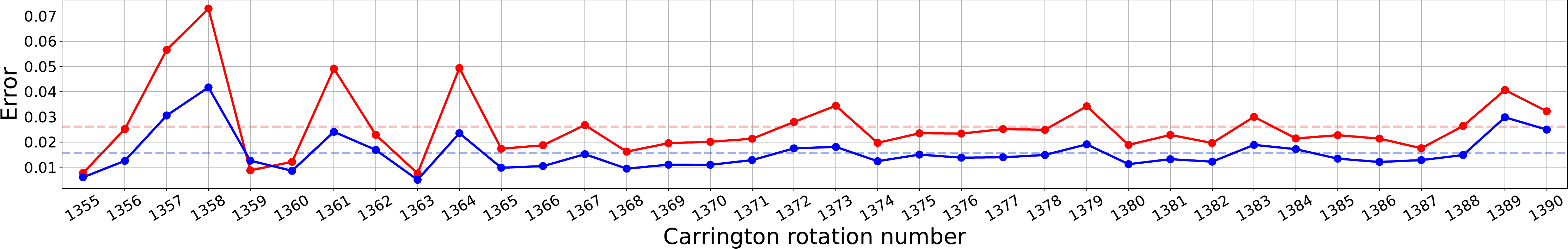}
        \caption{Error values $\E$ (in blue) and $\E^*$ (in red) for the whole considered time period. Their mean values are 0.016 and 0.026 and are shown with dashed horizontal lines.}
        \label{fig:errors_1}
    \end{figure}

    Of course, by increasing the number of neurons in the model, a perfect match between the target and predicted polarity maps can be achieved. For example, with enough neurons, the model can simply memorize the value for each pixel. However, it seems that such a model will be impractical in more realistic cases where the number of reference points on the synoptic map is smaller. It is therefore reasonable to prefer models with smaller numbers of neurons, to admit that the model may be imperfect in small-scale details, but to expect that such a model will be more stable.  Figure~\ref{fig:errors_1} shows that the errors for the proposed model architecture are
    consistent across the entire set of synoptic maps, with the majority being less than 5\%.
    
\subsection{Grid-Based Reference Points With Step 32}
    Now we consider a more realistic situation where the polarity is mostly
    unknown. We use step 32 for the reference points grid and train 100 independent
    models for each synoptic map. We use a larger number of independent models because the variability of the results increases due to a significant reduction in constraints.

\begin{figure}[h]
        \centering
        \includegraphics[width=\textwidth]{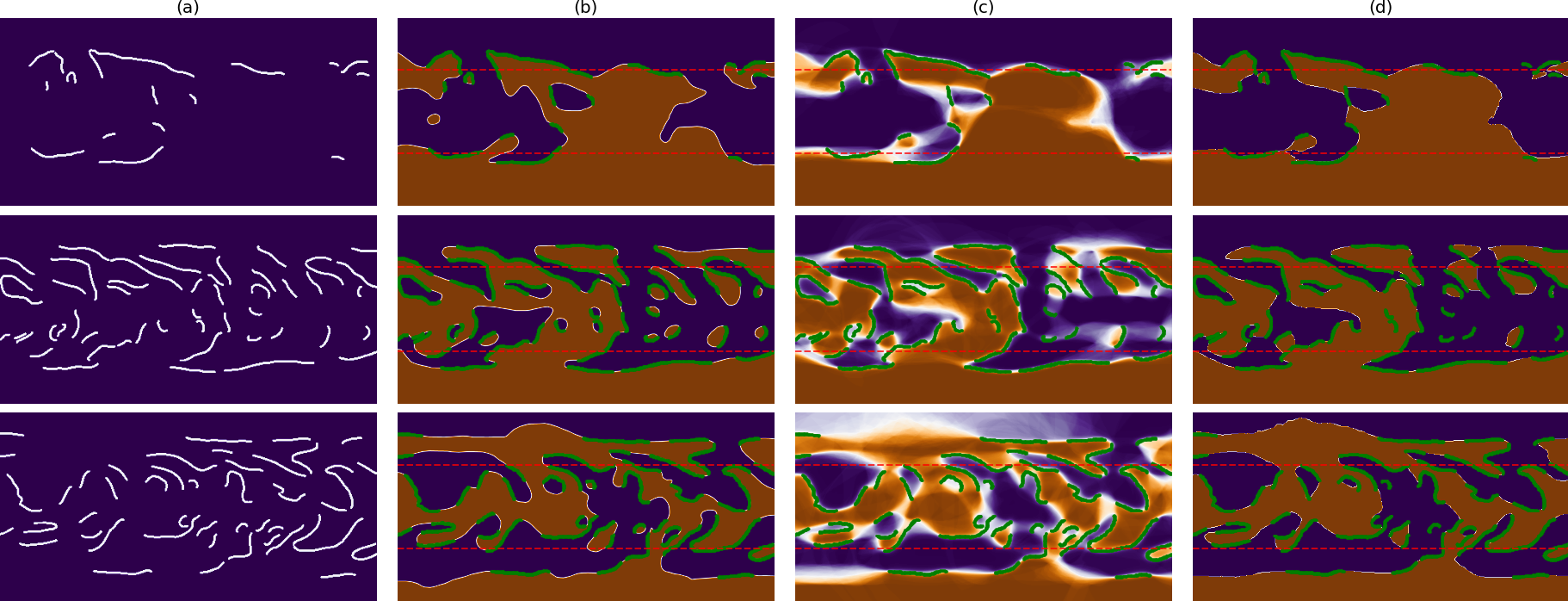}
        \caption{Reconstruction of polarity maps using a grid of reference points with a step 32 pixels.
        The plot legends are similar to Figure~\ref{fig:example_1}.}
        \label{fig:example_31}
    \end{figure}

    We observe in Figure~\ref{fig:example_31} that despite the significant
    reduction in the number of reference points, the model demonstrates robust performance.
    Considering the binarized maps, we can again note that some small scale structures are lost compared to the target maps, however, we can note that most of these regions in the averaged maps have low intensity (appear as white areas). This means that the model is not confident in the sign of these regions and the binarized values may be incorrect.

\begin{figure}[h]
        \centering
        \includegraphics[width=\textwidth]{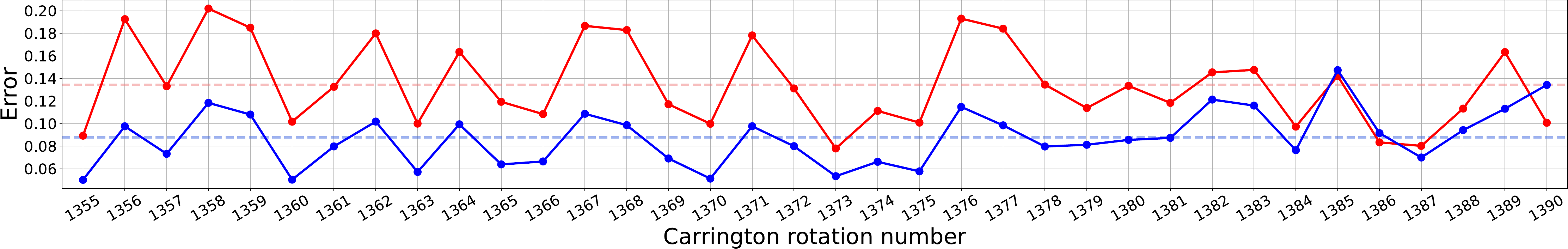}
        \caption{
        Error values $\E$ (in blue) and $\E^*$ (in red) for the whole considered time period. Their mean values are 0.088 and 0.135 and are shown with dashed horizontal lines.}
        \label{fig:errors_31}
    \end{figure}

    Figure~\ref{fig:errors_31} shows that the model accuracy is rather uniform across the full set of synoptic maps and the error rate fluctuates around 14\% for the entire map and 9\% for the low-latitude region. 
    
\begin{figure}[h]
        \centering
        \includegraphics[width=\textwidth]{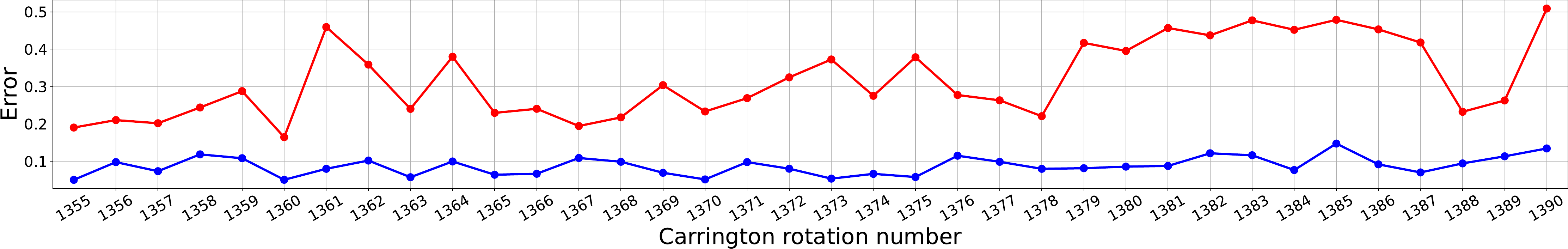}
        \caption{Error rates obtained for different strategies of aggregation of multiple predictions. The first strategy is to binarize the mean value (blue line), the second approach is to choose the prevailing sign for the binarized predictions (red line).}
        \label{fig:errors_strat}
    \end{figure}

    In this experiment, we also compare different strategies for aggregation of multiple predictions. Figure~\ref{fig:errors_strat} shows that binarization of the mean of the predicted values yields much lower errors than when we first apply binarization and then choose the prevailing sign.

\subsection{Grid-Based Reference Points With Step 64}
    In the next experiment we increase the grid step to 64. This means that we have prior information in only 32 pixels of the whole synoptic map of size $256\times 512$ pixels. Similar to the previous experiment, for each of the synoptic maps we train 100 independent models and average the results.

\begin{figure}[h]
        \centering
        \includegraphics[width=\textwidth]{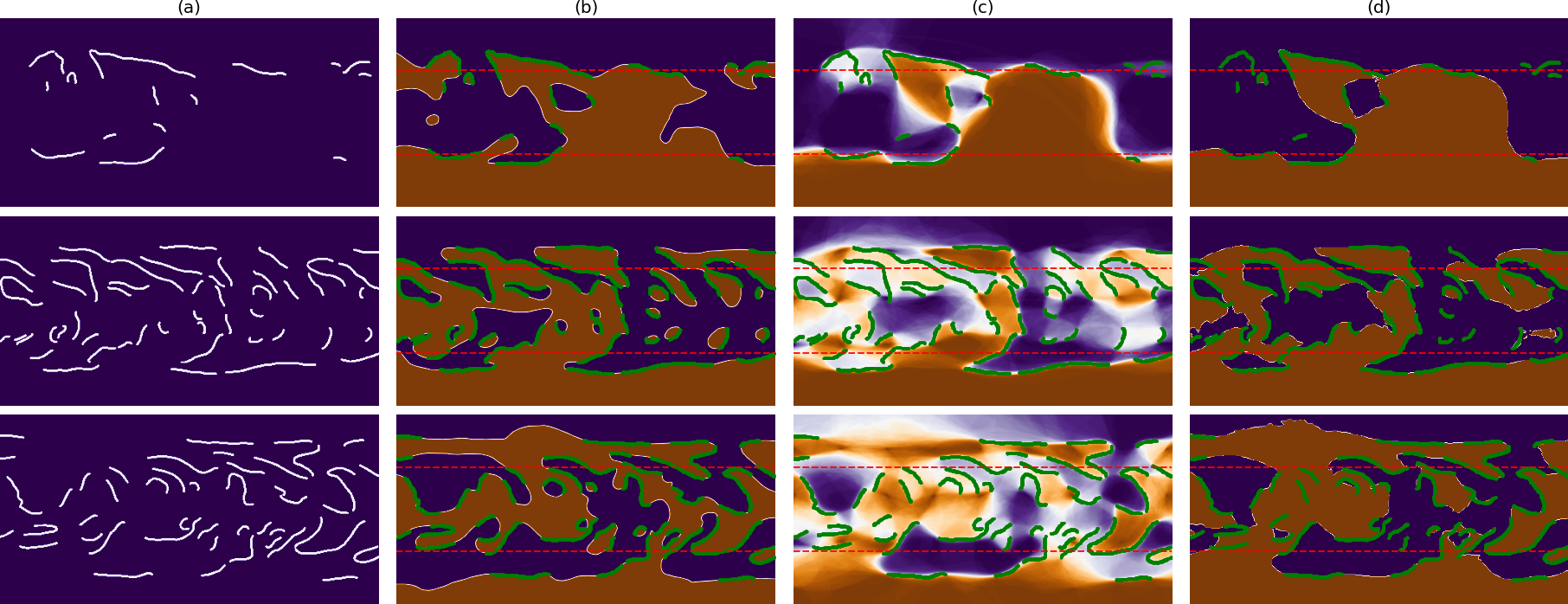}
        \caption{Reconstruction of polarity maps using a grid of reference points with a step 64 pixels.
        The plot legends are similar to Figure~\ref{fig:example_1}.}
        \label{fig:example_63}
    \end{figure}

\begin{figure}[h]
        \centering
        \includegraphics[width=\textwidth]{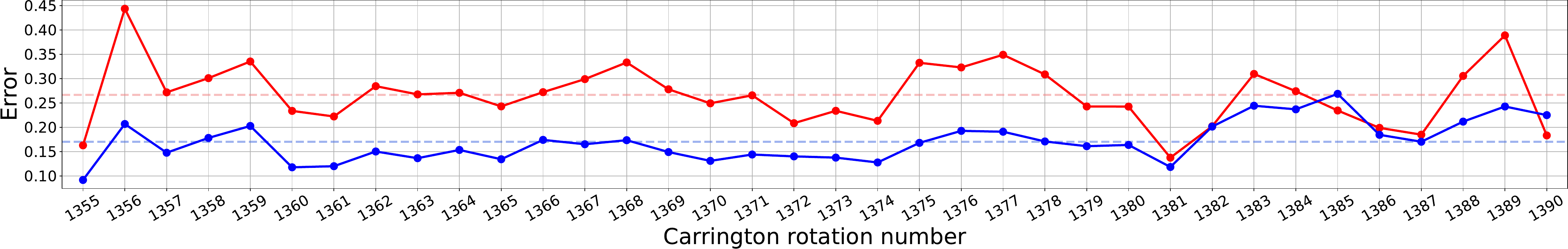}
        \caption{Error values $\E$ (in blue) and $\E^*$ (in red) for the whole considered time period. Their mean values are 0.170 and 0.267 and are shown with dashed horizontal lines.}
        \label{fig:errors_63}
    \end{figure}

    Figure~\ref{fig:example_63} shows that the averaged model prediction becomes more
    uncertain compared to the previous case with grid step 32, however, the averaged map
    still looks consistent with the position of filaments and target synoptic map.
    Due to the larger uncertainty (larger variation between independent 
    realizations of the polarity map) the binarized map becomes less 
    similar to the target map and, in particular, some filaments can be found
    in unipolar regions.
    Figure~\ref{fig:errors_63} shows that the average error rate increases up to
    27\% for the whole map and 17\% for the low-latitude region.

\subsection{Model Performance without Reference Points}
    Finally, we consider the reconstruction of magnetic polarity maps without reference points.
    This is the most challenging scenario for the model because it relies entirely on the filament data and 
    the loss function \(L(f)\) for training. 

\begin{figure}[h]
        \centering
        \includegraphics[width=\textwidth]{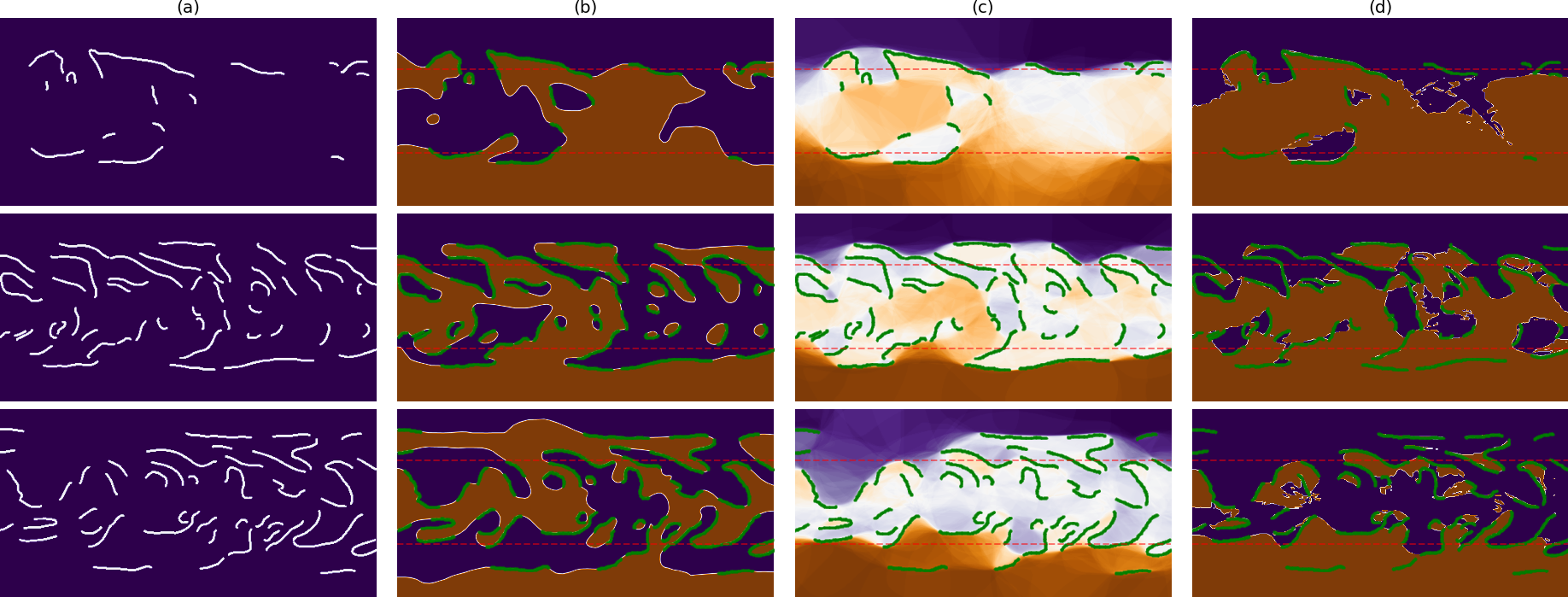}
        \caption{Reconstruction of polarity maps without using reference points.
        The plot legends are similar to Figure~\ref{fig:example_1}.}
        \label{fig:example_None}
    \end{figure}

    Figure~\ref{fig:example_None} shows the reconstruction results for several synoptic maps.
    We observe that a large region of the synoptic map has low confidence, and a mostly inconsistent map
    is obtained after binarization of the averaged map.
    
    As expected, the absence of reference points leads to a decrease in the accuracy of the reconstructed polarity maps. The average error rates for the entire map and the low-latitude region are approximately \(0.35\) and \(0.5\), respectively.
    However, the results of this experiment are important for understanding the limits of the predictive ability of the model when it operates under minimal information constraints. 
    We attribute the low accuracy of these results to the fact that the loss function \(L(f)\) 
    admits solutions in which the filaments do not separate opposite polarities.

\section{Conclusions and Discussion}
    We proposed a new approach based on machine learning to reconstruct
    magnetic polarity maps from synoptic maps of filaments.
    The main feature of this approach is that it does not require training data sets.
    Instead, for each synoptic map we formulate a loss function 
    based on the filament data and general physical constraints, and optimize the 
    network model to minimize this loss function.
    This approach can be compared to a class of physics-informed neural network models \citep[PINN, see, e.g.,][]{Raissi, Baty, Jarolim}.
    The difference is that instead of using an explicit partial differential equation,  
    we use a set of weaker constraints.

    To construct a function that satisfies the required constraints, we used a rather simple neural network model with a fully-connected architecture. A typical drawback of fully-connected models is that they contain a large number of neurons (trainable parameters). In our case, the maximum number of neurons in a layer is 24, and the model has a total of 823 trainable parameters. For comparison, the fully-connected model used in \cite{Jarolim_2024} for magnetic-field reconstruction consists of layers with 256 neurons, and the number of trainable parameters is of the order of the square of the number of neurons.

    We appreciate that the class of neural network architectures that can be used in our case is much broader than just fully-connected models. For example, convolutional neural network (CNN) models are typical in the context of image processing. In particular, the U-Net \citep{unet} model looks suitable for image-to-image problems. This choice does not reduce the number of neurons (the original implementation has more than 1M trainable parameters). More importantly, the convolutional model relies essentially on the 2D geometry of the input data. In our approach, we use 3D geometry that considers the distance between points on the synoptic map not only along the solar surface but also in 3D. Moreover, the reconstructed polarity map can be naturally extrapolated to the solar interior and corona, whereas this is not straightforward for 2D convolutional models.

    In terms of computational performance, the simplicity of the model is compensated by the large number (30K) of training iterations we use to train the model. As a result, training each model takes 3--4 minutes on a GPU laptop. Since we train 100 independent models for each synoptic map, the total training time is about 6 hours. Of course, this time can be reduced by training models in parallel.

    It is also possible to formulate the problem of magnetic polarity reconstruction
    as a supervised machine-learning problem. For example, one could train a CNN model or a generative adversarial network (GAN) model
    using a set of synoptic maps with filaments as inputs and polarity maps as targets
    \citep[see, e.g.,][for more details on these techniques]{Ramos, Camporeale}.
    Although this approach looks attractive, we assume that the number of available 
    synoptic maps is not sufficient for training. One more problem
    with such an approach is that the targets are not the ground truth, but the results of the 
    reconstruction results and contain uncertainties.
    
    Of course, our approach is also not perfect. The main issue is that the without any prior or additional information, the problem of polarity
    reconstruction from filaments position allows too much variability in the results. The situation can be improved in several ways.
    The first one is introducing additional terms in the loss function
    that will constrain the results. However, it is not obvious how to
    formulate such constraints  explicitly. 
    Probably, at least some of them could be derived from topological
    considerations \citep[see, e.g.,][]{Makarenko}, from
    dynamo theory \citep[see, e.g.,][]{Obridko}, or from long-term properties
    of photospheric magnetic fields \citep[e.g.,][]{Pevtsov}.
    The second way is to use information from the previous or next synoptic maps
    and require that the difference between two neighboring synoptic maps can not be
    dramatic. In particular, this can be implemented by initializing the neural 
    network model not with random weights, but with weights learned for the previous polarity map.
    
    In our research, we compensate for the lack of proper constraints by 
    assuming that the polarity is known for some points on the synoptic map.
    It was demonstrated that a grid of reference points with step 
    1/8 of the vertical size of the synoptic map
    together with positions of filaments yields a reconstruction visually close
    to the reconstruction proposed by McIntosh. The number of incorrectly classified
    pixels is below 10\% for the region $\pm 40^{\circ}$ from the solar equator.
    Increasing the step size we noted that the averaged 
    prediction becomes more uncertain and may become locally inconsistent
    with the filament data after binarization.

\begin{figure}[h]
        \centering
\includegraphics[width=\textwidth]{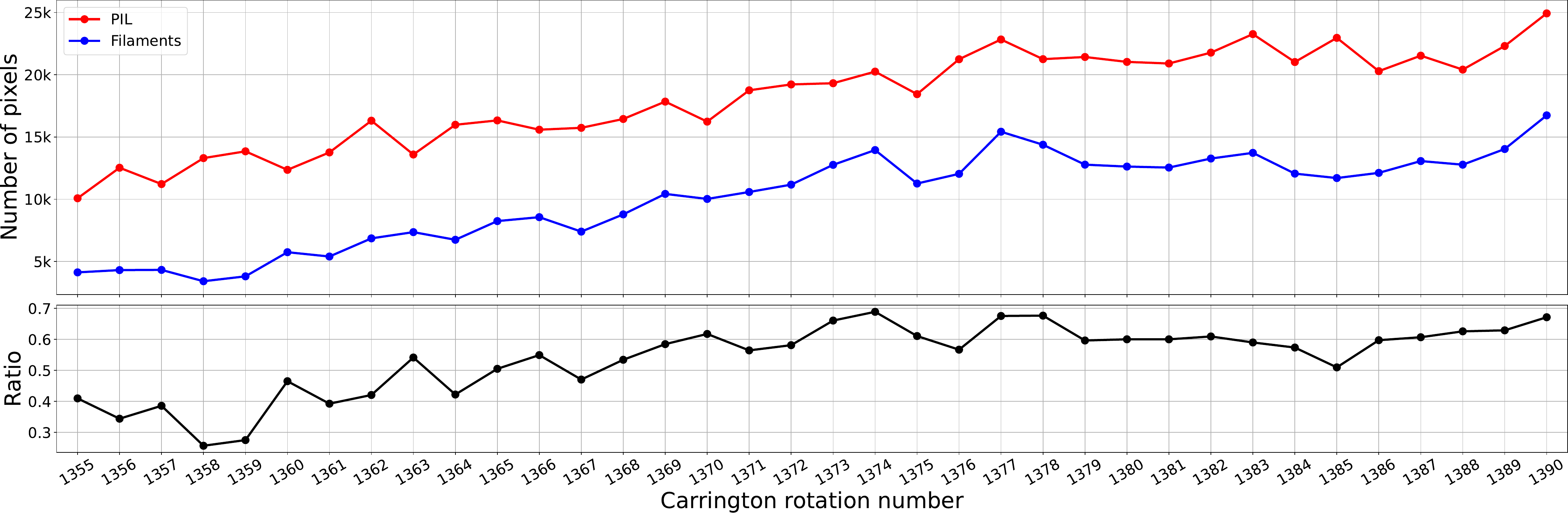}
        \caption{Upper panel: total number of pixels that belong to filaments (the blue line) and polarity inversion lines (the red line) in a synoptic map of size $512\times256$ pixels. The bottom panel shows the ratio of these numbers.}
        \label{fig:filaments_lines}
    \end{figure}

    One more reason for the uncertain predictions may be the complexity of the filament structure. As a simple and accessible test, we compared the total length of the filaments, the total length of the PILs, and the error rate of the reconstructed polarity map (more precisely, we compute the total length as the total number of pixels that belong to filaments and PILs). Figure~\ref{fig:filaments_lines} shows that the total length of the filaments and the PILs as well as their ratio increase with time. However, as we observed in Figures~\ref{fig:errors_1}, \ref{fig:errors_31}, and \ref{fig:errors_63}, the error rate is rather stable over time. More formally, the correlation coefficient between the ratio of total length of filaments to total length of PILs and the error rate in the reconstructed polarity maps is only about 0.2. Probably, more complex topological parameters of the polarity map could better explain the errors, but we leave it for separate research.

    We consider this research as an attempt to understand what 
    information and in what form should be provided to machine-learning models 
    to reconstruct magnetic polarity maps and how to define what is a correct reconstruction.
    We proposed a way in which we see a possible solution and made the first steps. But even now the model could be useful for  
    semi-supervised reconstruction of polarity maps. In this process, the user
    defines polarity for some points and the model proposes the reconstruction
    most consistent with filaments and this additional information. The process can be iterative, so the user can start with a few points and
    then refine the results by adding more points where necessary.
    Another source of semi-supervised information could be low-resolution or local magnetic field measurements, and thus our model can be considered as a super-resolution tool that
    combines magnetic and chromospheric data.
    
    The goal of the future research is to minimize the dependence on the
    auxiliary information and to create a long-term, homogeneous,
    and reproducible series of magnetic polarity maps.
    
    Finally, we note that the source code of the model, reconstruction process
    and full set of reconstructed polarity maps are available in GitHub
    repository \href{https://github.com/observethesun/PIL}{github.com/observethesun/PIL}.

\begin{acks}
    EI acknowledges the support of RSF grant 21-72-20067.
\end{acks}

\section*{Data availability}
    The code and datasets generated in this study are available in the GitHub repository \href{https://github.com/observethesun/PIL}{github.com/observethesun/PIL}.

\begin{conflict}
    The authors declare that they have no conflicts of interest.
\end{conflict}

\bibliographystyle{spr-mp-sola}
\bibliography{literature}

\end{document}